%% file: submission.tex
\documentclass[10pt,twocolumn,letterpaper]{article}

\usepackage{iccv}
\usepackage{booktabs}
\usepackage{times}
\usepackage{epsfig}
\usepackage{graphicx}
\usepackage{amsmath}
\usepackage{amssymb}
\usepackage{multirow}
\usepackage{comment}

\input{math_commands.tex}

\usepackage[utf8]{inputenc} %
\usepackage[T1]{fontenc}    %
\usepackage{url}            %
\usepackage{amsfonts}       %
\usepackage{nicefrac}       %
\usepackage{microtype}      %
\usepackage{xcolor}         %
\usepackage[normalem]{ulem}
\usepackage{wrapfig}

\usepackage{subcaption}     %
\usepackage{xspace}         %
\usepackage{siunitx}       %
\usepackage{authblk}       %

\newcommand{\ourmodel}{HVG\xspace}
\newcommand{\dvdgan}{DVD-GAN\xspace}

\usepackage[pagebackref=true,breaklinks=true,letterpaper=true,colorlinks,bookmarks=false]{hyperref}

\iccvfinalcopy %

\def\iccvPaperID{10501} %

\ificcvfinal\fi

\begin{document}

\title{Hierarchical Video Generation for Complex Data}
\author[1, 2]{Lluis Castrejon}
\author[2]{Nicolas Ballas}
\author[1, 3]{Aaron Courville}

\affil[1]{Mila, Université de Montréal}
\affil[2]{Facebook AI Research}
\affil[3]{CIFAR Fellow}

\vspace{-3ex}
\maketitle

\ificcvfinal\thispagestyle{empty}\fi

\input{abstract}

\input{intro_v3}

\input{method_v2}

\input{related_work}
\input{experiments_v3}
\input{conclusions}

\section*{Acknowledgements}
This research was sponsored by a CIFAR Canada AI Chair given to AC and by a IVADO PhD Fellowship given to LC.
We thank Pascal Vincent, Quentin Duval and Mike Rabbat for their feedback.

{\small
\bibliographystyle{ieee_fullname}
\bibliography{bibliography}
}

 \input{appendix}

\end{document}

%% file: math_commands.tex
\usepackage{amsmath,amsfonts,bm}

\def\eqref#1{equation~\ref{#1}}

\def\1{\bm{1}}

\DeclareMathAlphabet{\mathsfit}{\encodingdefault}{\sfdefault}{m}{sl}
\SetMathAlphabet{\mathsfit}{bold}{\encodingdefault}{\sfdefault}{bx}{n}

%% file: abstract.tex
\begin{abstract}

Videos can often be created by first outlining a global description of the scene and then adding local details.
Inspired by this we propose a hierarchical model for video generation which follows a coarse to fine approach. 
First our model generates a low resolution video, establishing the global scene structure, that is then refined by subsequent levels in the hierarchy.
We train each level in our hierarchy sequentially on partial views of the videos.
This reduces the computational complexity of our generative model, which scales to high-resolution videos beyond a few frames. 
We validate our approach on Kinetics-600 and BDD100K, for which we train a three level model capable of generating 256x256 videos with 48 frames.

\end{abstract}

%% file: intro_v3.tex
\section{Introduction}

Humans have the ability to simulate novel visual objects and their dynamics using their imagination. 
This ability is linked to the capacity to perform planning or counter-factual thinking.  
Replicating this ability in machines is a longstanding challenge that generative models address.
Advances in generative modeling and increased computational resources have enabled the generation of realistic high-resolution images~\cite{brock2018large} or coherent text in  documents~\cite{brown2020language}. 
Yet, models for video generation have been less successful, in part due to their high memory requirements that scale with the generation resolution and length.

\begin{figure}[t]
    \centering
    \includegraphics[trim=20 10 20 10, width=\linewidth, clip]{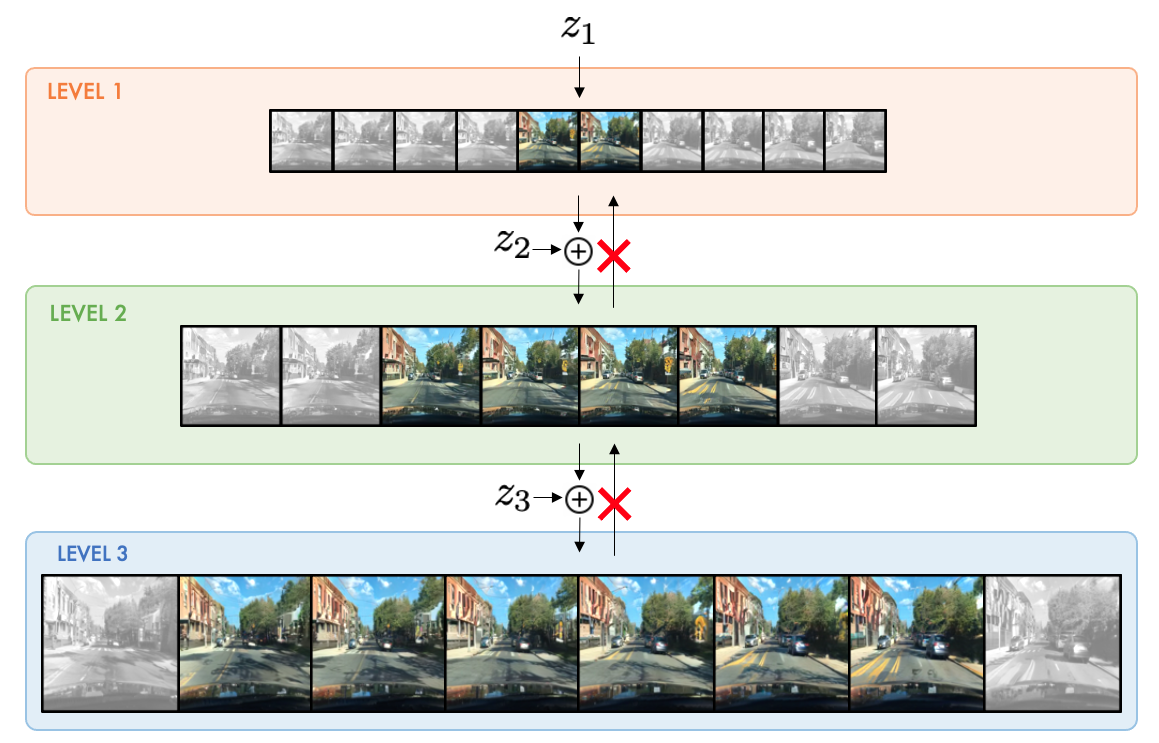}
    \caption{\small \textbf{Hierarchical Video Generation}
        We propose to divide the generative process into multiple simpler problems.
        \ourmodel first generates a low resolution video that depicts a full scene at a reduced framerate. 
        This scene outline is then progressively upscaled and temporally interpolated.
        Levels are trained sequentially and do not backprogagate the gradient to the previous levels. Additionally, upscaling levels are trained on temporal crops of their inputs during training (illustrated by the non-shaded images) to reduce their computational requirements. 
        Our model is competitive with the state-of-the-art in video generation and enables the generation of longer high resolution videos than possible with previous methods.
    }
    \label{fig:teaser}
\end{figure}

When creating visual data, artists often first create a rough outline of the scene, to which then they add local details in one or multiple iterations.
The outline ensures global consistency of the scene and divides the creative process into multiple tractable steps.
Inspired by this strategy we propose \ourmodel, a hierarchical video generation model which divides the generative process into a set of simpler problems.
\ourmodel first generates a low-resolution video that depicts a full scene at a reduced framerate. 
This scene outline is then progressively upscaled and temporally interpolated to the desired  resolution by one or more upscaling levels as depicted in Figure~\ref{fig:teaser}.
Every level in our model hierarchy outputs a video that serves as the input to the next one, with each level specializing in a particular aspect of the generation.

Levels in our hierarchy are trained greedily, i.e.\ in sequence and not end-to-end. 
This sequential training allows us to consider only one hierarchy level at a time and thus reduce the training memory requirements.
We formulate each level as a adversarial game that we solve leveraging the GAN framework. 
We theoretically demonstrate that our training setup admits the same global solution as an end-to-end model.

While the first level produces a complete but low-resolution video, the upscaling levels are applied only on temporal crops of their inputs during training to reduce their computational requirements.
Despite their temporally-local training,  the upscaling levels of the hierarchy are capable of producing long videos with temporal coherence at inference due to their conditioning on the global scene outline.
This enables \ourmodel to generate higher-resolution videos with a larger number of frames than is possible with previous methods. 
Our contributions can be summarized as follows:
\begin{itemize}
\item We define a hierarchical approach for video generation which divides the generation process into multiple tractable steps. 
\item We empirically validate our approach on Kinetics-600 and BDD100K, two large-scale datasets with complex videos in real-world scenarios. \ourmodel is competitive with \dvdgan, the state-of-art generative model of videos~\cite{clark2019efficient}.
In particular, our approach
outperforms DVD-GAN model in FID score and reaches a similar IS score when generating videos of 48 frames at resolution 128x128 on the Kinetics-600 dataset.

\item We use our approach to generate videos with 48 frames at a resolution of 256x256 pixels, beyond what is currently possible with other methods. 
To the best of our knowledge, our method is the first video generative model to produce such large generations. 
\end{itemize}

%% file: method_v2.tex
\section{Hierarchical Video Generation}
\label{section:method}

Video scenes can be created by first outlining the global scene and then adding local details.
Following this intuition we propose \ourmodel, a hierarchical model in which each stage only considers a lower dimensional view of the data, first at a global level and then locally. 
Video generation models struggle to scale to high frame resolutions and long temporal ranges.
The goal of our method is to break down the generation process into smaller steps which require less computational resources when considered independently.

\paragraph{Problem Setting}

We consider a dataset of videos $(\mathbf{x}_1, ..., \mathbf{x}_n)$ where each video $\mathbf{x}_i = (\mathbf{x}_{i; 0}, ... , \mathbf{x}_{i; T})$ is a sequence of $T$ frames $\mathbf{x}_{i; t} \in \mathbb{R}^{H\times W\times 3}$.
Let $f_s$ denote a spatial bilinear downsampling operator and  $f_t$ a temporal subsampling operator. 
For each video $\mathbf{x}_i$, we can obtain lower resolution views of our video by repeated application of $f_s$ and $f_t$, i.e. $\mathbf{x}^l_{i} = f_s(f_t(\mathbf{x}^{l+1}_{i}), \forall l \in [1..L]$ with  $\mathbf{x}^L_{i}=\mathbf{x}_{i}$. 

Each ($\mathbf{x}^1_{i}$, ..., $\mathbf{x}^L_{i}$) comes from a joint data distribution $p_{d}$. 
Our goal is to learn a generative distribution $p_g$ such that $p_g=p_{d}$.

\paragraph{Hierarchical Generative Model}
 
We define a generative model that approximates the joint data distribution according to the following factorization:
\begin{eqnarray}
p_{g}(\mathbf{x}^1, ..., \mathbf{x}^L) =  p_{g_{L}}(\mathbf{x}^L | \mathbf{x}^{L-1}) ... p_{g_2}(\mathbf{x}^2 | \mathbf{x}^1) p_{g_1}(\mathbf{x}^1).
\label{eq:gen_f}
\end{eqnarray}
Each $p_{g_i}$ defines a level in our model. 
This formulation allows us to decompose the generative process in to a set of smaller problems.
The first hierarchy level $p_{g_1}$ produces low resolution and temporally subsampled videos from a latent variable. 
For subsampling factors $K_{T}$ and $K_{S}$ (for time and space respectively), the initial level generates videos $\mathbf{x}^{1}_i = (\mathbf{x}^{1}_{i; 0}, \mathbf{x}^{1}_{i; K_{T}}, \mathbf{x}^{1}_{i; 2K_{T}}, ... , \mathbf{x}^{1}_{i; T})$, which is a sequence of $\frac{T}{K_{T}}$ frames $\mathbf{x}^{1}_{i; t} \in \mathbb{R}^{\frac{H}{K_{S}} \times \frac{W}{K_{S}} \times 3}$. 
The output of the first level is spatially upscaled and temporally interpolated by one or more subsequent upscaling levels.

\paragraph{Training}
We train our model one level at a time and in order, i.e.\ we train the first level to generate global but downscaled videos, and then we train upscaling stages on previous level outputs one after each other.
We do not train the levels in an end-to-end fashion, which allows us to break down the computation into tractable steps by only training one level at a time.
We formulate a GAN objective for each level in the hierarchy.
We consider the distribution $p_{g_1}$ in eq.~\ref{eq:gen_f} and solve a min-max game with the following value function:
{\small
\begin{eqnarray}
\mathbb{E}_{\mathbf{x}^1\sim p_{d}} [\log (D_1(\mathbf{x}^1))] + \mathbb{E}_{\mathbf{z}_1 \sim p_{z_1}} [\log (1-D_1(G_1(\mathbf{z}_1)))],
\label{eq:gan_stage1}
\end{eqnarray}} where $G_1$ and $D_1$ are the generator/discriminator associated with the first stage and $p_{z_1}$ is a noise distribution.
This is the standard GAN objective~\cite{goodfellow2014generative}. 
For upscaling levels corresponding to $p_{g_l}, l > 1$, we consider the following value function:
{\small
\begin{eqnarray}
&  \mathbb{E}_{\mathbf{x}^{l-1}, ..., \mathbf{x}^{1}\sim p_{d}} \mathbb{E}_{\mathbf{x}^l\sim p_{d}( .| \mathbf{x}^{l-1}, ..., \mathbf{x}^{1})} [\log(D_l(\mathbf{x}^{l}, \mathbf{x}^{l-1}))] + \nonumber \\ & \mathbb{E}_{\mathbf{\hat{x}}^{l-1}\sim p_{g_{l-1}}} \mathbb{E}_{\mathbf{z}_l\sim p_{z_l}} [\log(1-D_l(G_l(\mathbf{z}_l, \mathbf{\hat{x}}^{l-1}),  \mathbf{\hat{x}}^{l-1}))],
\label{eq:gan_stage2}
\end{eqnarray}} where $G_l$, $D_l$ are the generator and discriminator of the current level and $p_{g_{l-1}}$ is the generative distribution of the level $l-1$.\footnote{We simplify the notation and denote by $p_{g_{l-1}}(\mathbf{x^{l-1}})$ the joint distribution $p_{g_{l-1}}(\mathbf{x}^{l-1}|\mathbf{x}^{l-1})... p_{g_2}(\mathbf{x}^2 | \mathbf{x}^1) p_{g_1}(\mathbf{x}^1)$.}
The min-max game associated with this value function has a global minimum when the two joint distributions are equal, $p_{d}(\mathbf{x}, ..., \mathbf{x^l}) = p_{g_l}(\mathbf{x^l} | \mathbf{x}^{l-1}) .. p_{g_1}(\mathbf{x}^1)$~\cite{dumoulin2016adversarially, donahue2016adversarial}.
For more details about the training objectives for each level and a complete proof that the sequential training admits the same global minimum than an end-to-end training refer to the Appendix.
We also see from eq.~\ref{eq:gan_stage2} that the discriminator for upscaling stages operates on pairs $(\mathbf{x}^l, \mathbf{x}^{l-1})$ videos to determine whether they are real or fake. 
This ensures that the upscaling stages are grounded on their inputs, i.e that $\mathbf{x}^l$ "matches" its corresponding $ \mathbf{x}^{l-1}$ .

\paragraph{Partial View Training}
Computational requirements for upscaling levels can be high when generating large outputs, even if each level is trained sequentially.
As we increase the length and resolution of a generation, models need to store large activation tensors during training that use large amounts of GPU memory.
To reduce the computational requirements, we apply the upscaling levels only on temporal crops of their inputs during training.
This strategy reduces training costs since we upscale smaller tensors, at the expense of having less available context to interpolate frames.

\paragraph{Inference}
At inference time, \ourmodel generates videos by applying the upscaling levels on their inputs without any temporal cropping.
Upscaling levels are convolutional, i.e.\ they learn functions that are applied in a sliding window manner over their inputs.
Since we do not temporally crop the inputs at inference time, the upscaling function is applied to all possible input windows to generate a full video.

\section{Model Parametrization}
\label{section:model_parametrization}

\begin{figure}
    \centering
    \includegraphics[trim={0 0 0 0},clip,width=\linewidth]{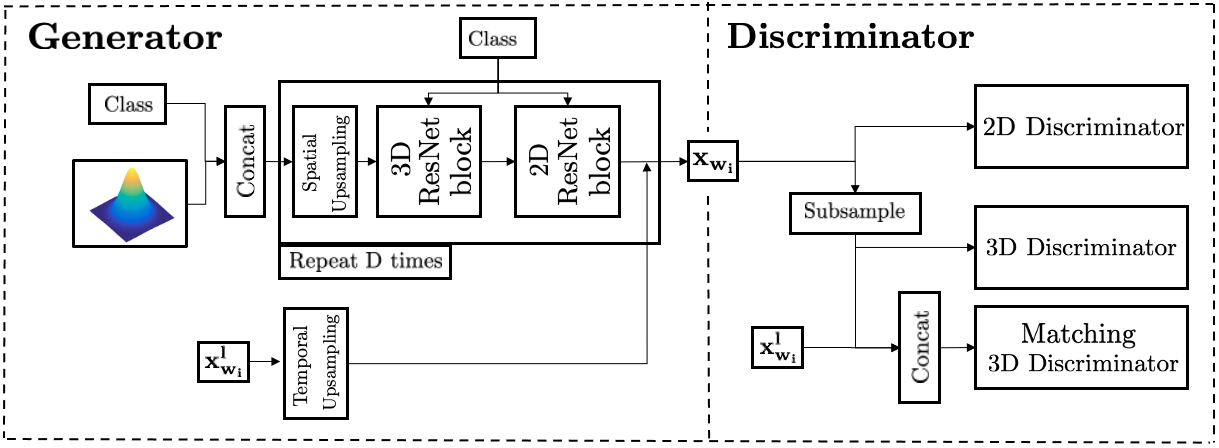}
    \caption{\small \textbf{Upsampling level parametrization}
    The upsampling levels use a conditional generator and three discriminators - spatial/2D, temporal/3D and matching. The conditional generator learns to upsample the previous level output, while the matching discriminator is trained on pairs of real/generated conditions and outputs.
    }
    \label{fig:upsampling_model}
\end{figure}

In this section we describe the parametrization of the different levels of \ourmodel.
We keep the discussion at a high level, briefly mentioning the main components of our model.
Precise details on the architecture are provided in the Appendix.
 
\paragraph{First Level} 
The first level generator stacks units composed by a ConvGRU layer~\cite{ballas2015delving}, modeling the temporal information, and 2D-ResNet blocks that upsample the spatial resolution.
Similar to MoCoGAN~\cite{tulyakov2018mocogan} and \dvdgan, we use a dual discriminator with both a spatial discriminator that randomly samples $k$ full-resolution frames and discriminates them individually, and a temporal discriminator that processes spatially downsampled but full-length videos.
The size of the latent variables and the number of channels used in G and D are described in the Appendix.

\paragraph{Upsampling Levels} 
The upsampling levels are composed by a conditional generator and three discriminators (spatial, temporal and matching). 
The conditional generator produces an upscaled version $\mathbf{\hat{x}^l}$ of a lower resolution video $\mathbf{\hat{x}^{l-1}}$.
To discriminate samples from real videos, upscaling stages use a spatial and temporal discriminator, as in the first level.
Additionally, we introduce a matching discriminator.
The goal of the matching discriminator is to ensure that the output is a valid upsampling of the input.
Without this discriminator, the upsampling generator could learn to ignore the low resolution input video.
The conditional generator is trained jointly with the spatial, temporal and matching discriminators.
Fig.~\ref{fig:upsampling_model} shows an overview of the upsampling level parametrization. 

\paragraph{Conditional Generator} 
The conditional generator takes as input a lower resolution video $\mathbf{\hat{x}}^{l-1}$, a noise vector $\mathbf{z}$ and optionally a class label $y$, and generates $\mathbf{\hat{x}}^{l}$.
We increase the duration of the low resolution video $\mathbf{\hat{x}}^{l}$ to the same temporal duration as the output by repeating its frames before feeding it to the generator as a condition.   

Our conditional generator stacks units composed by one 3D-ResNet block and two 2D-ResNet blocks. 
Spatial upsampling is performed gradually by progressively increasing the resolution of the generator blocks.
To condition the generator we add residual connections~\cite{he2016deep, srivastava2015highway} from the low-resolution video to the output of each generator unit.
We sum nearest-neighbor interpolations of the lower resolution input to each unit output.
We do not use skip connections for units whose outputs have higher spatial resolution than the lower dimensional video input, i.e.\ we do not upscale the low resolution video.

\paragraph{Matching Discriminator} 
The matching discriminator uses an architecture like that of the temporal discriminator. 
It discriminates real or generated input-output pairs.
The output is downsampled to the same size as the input, and both tensors are concatenated on the channel dimension.
A precise description of all discriminator architectures can be found in the Appendix.

%% file: related_work.tex
\section{Related Work}

Since~\cite{ranzato2014video} and~\cite{video_lstm} proposed the first video generation models inspired by the language modeling literature, many papers have proposed different approaches to represent and generate videos~\cite{luc2017predicting, luc2018predicting,villegas2017decomposing, villegas2017learning,xue2016visual}.

Autoregressive models~\cite{larochelle2011neural,dinh2016density,kalchbrenner2017video,reed2017parallel,weissenborn2019scaling} model the conditional probability of each pixel value given the previous ones.
They do not use latent variables and their training can be easily parallelized.
Inference in autoregressive models often requires a full forward pass for each output pixel, which does not scale well to long high resolution videos whose output dimensionality can be in the million pixels.

Normalizing flows~\cite{rezende2015variational,kingma2018glow,kumar2019videoflow} learn bijective functions that transform latent variables into data samples.
By defining bijective functions normalizing flows are able to directly maximize the data likelihood, which eases their training.
However they require the latent variable to have the same dimensionality as its output, which becomes an obstacle when generating videos due to their large dimensionality.

Variational AutoEncoders (VAEs)~\cite{kingma2013auto, rezende2014stochastic, sv2p} also transform latent variables into data samples.
While more scalable, VAEs often produce blurry results when compared to other generative models.
Models based on VRNNs~\cite{vrnn, svg, savp, castrejon2019improved} use one latent variable per video frame and often produce better results.

Generative Adversarial Networks (GANs) are another type of latent variable models which optimize a min-max game between a generator G and a discriminator D trained to tell real and generated data apart~\cite{goodfellow2014generative}.
Empirically, GANs usually produce better samples than competing approaches but might suffer from mode collapse.
GAN models for video were first proposed in~\cite{vondrick2016generating, vondrick2016anticipating, mathieu2015deep}.
In recent work, SAVP~\cite{savp} proposed to use the VAE-GAN~\cite{larsen2015autoencoding} framework for video. 
TGANv2~\cite{saito2018tganv2} improves upon TGAN~\cite{TGAN2017} and proposes a video GAN trained on data windows, similar to our approach. 
However, unlike TGANv2, our model is composed of multiple stages which are not trained jointly.
MoCoGAN~\cite{tulyakov2018mocogan} first introduced a dual discriminator architecture for video, with \dvdgan~\cite{clark2019efficient} scaling up this approach to high resolution videos in the wild.
\dvdgan outperforms MoCoGAN and TGANv2, and is arguably the current state-of-the-art in adversarial video generation.

Our model is also related to work that proposes hierarchical or progressive training approaches for generative models~\cite{karras2017progressive,denton2015deep,xiong2018learning,zhao2020towards}. 

%% file: experiments_v3.tex
\section{Experiments}

\begin{figure*}[!ht]
    \centering
    \includegraphics[width=\textwidth]{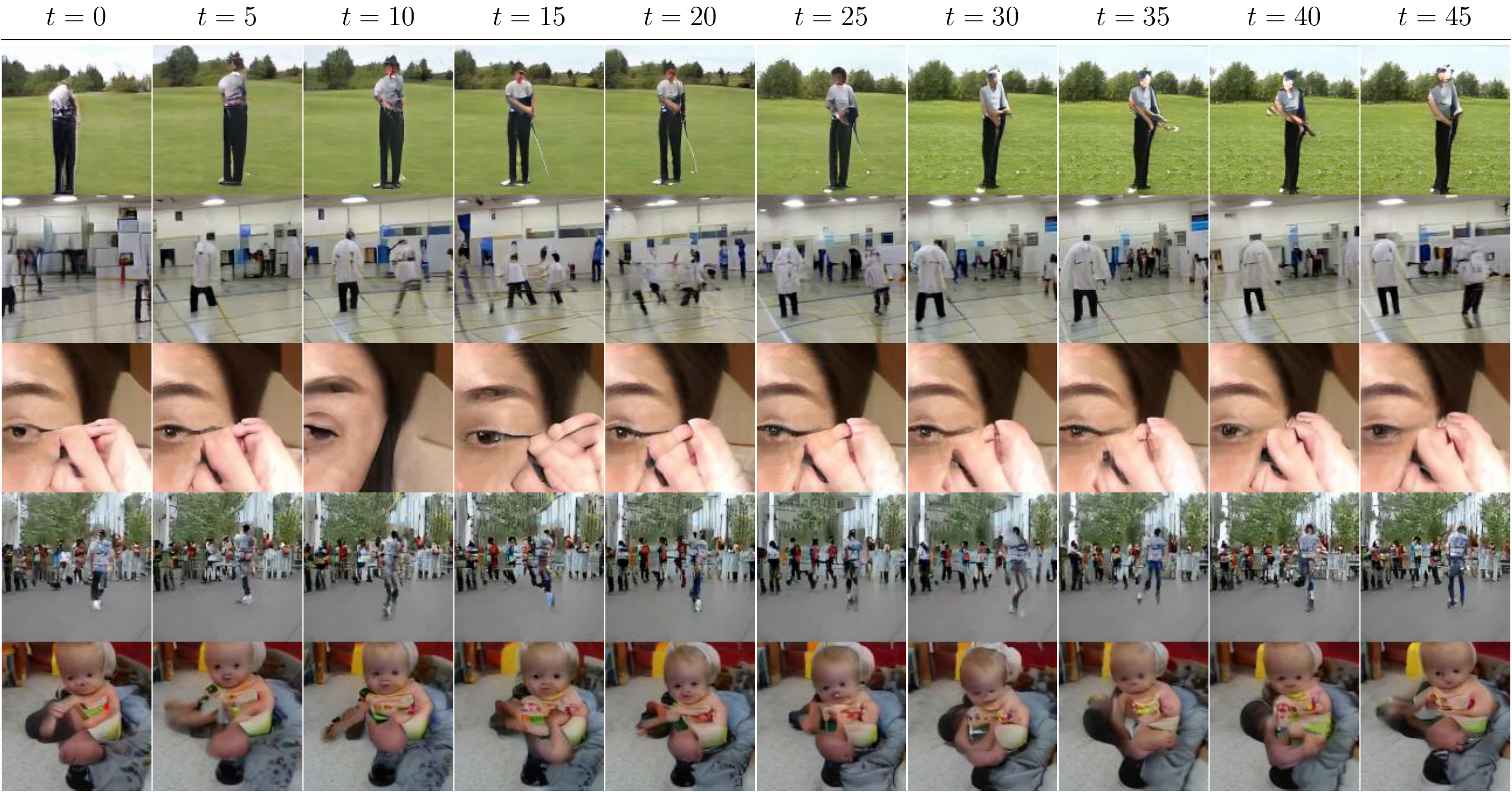}
   \vspace{-1em}
    \caption{
    \small \textbf{Randomly selected \ourmodel 48/128x128 frame samples for Kinetics-600:} 
    These samples were generated by unrolling \ourmodel 12/128x128 to generate 48 frame sequences, 4 times its training horizon. 
    Each row shows frames from the same sample at different timesteps.
    The generations are temporally consistent and the frame quality does not degrade over time.
    }
    \label{fig:samples_kinetics_128}
\end{figure*}

\begin{table*}
    \centering
    \begin{tabular}{lccccccc}
    \toprule
    & & \multicolumn{3}{c}{Evaluated on 12 frames} & \multicolumn{3}{c}{Evaluated on 48 frames}\\
    \cmidrule(lr){3-5} \cmidrule(lr){6-8} 
    Model & Trained on & IS ($\uparrow$) & FID ($\downarrow$) & FVD ($\downarrow$) & IS ($\uparrow$) & FID ($\downarrow$)  & FVD ($\downarrow$) \\
    \midrule
          DVD-GAN & 12/128x128 & 77.45 & \textbf{1.16} & - & N/A & N/A & N/A \\
          DVD-GAN & 48/128x128 & N/A & N/A & N/A & \textbf{81.41} & 28.44 & -  \\
    \midrule
          2-Level \ourmodel & 12/128x128 & \textbf{104.00} & 2.09 & 591.90 & 77.36 & \textbf{14.00} & 517.21 \\
    \bottomrule
    \end{tabular}
    \caption{\small \textbf{Results on Kinetics-600 128x128} We compare our two-level \ourmodel model against the reported metrics for two \dvdgan models~\cite{clark2019efficient}, one trained to generate 12 frames and one trained to generate 48 frames. Our model is trained to only generate 12 frames and is able to match the performance of the 12-frame \dvdgan model.
    Additionally, the same \ourmodel model is able to generate 48 frames when applied convolutionally over the full first level output.
    In that setup our model also matches the quality of a 128x128 \dvdgan model trained to generate 48 frames, while having computational requirements close to the 12/128x128 version.
    }
    \label{tab:dvdgan_comparison}
\end{table*}

In this section we empirically validate our proposed approach.
First, we show in Section~\ref{sec:baseline_comparison} that our approach is competitive with the state-of-the-art performance on Kinetics-600 for 128x128 videos.
Then, we analyze the scaling properties of our model to generate longer higher resolution videos in Section~\ref{sec:scaling}.
Finally, we provide an ablation of the main components of our model in Section~\ref{sec:ablations}.

For the rest of the section we denote video dimensions by their output resolution DxD and number of frames F in the format F/DxD.

\subsection{Experimental Setting}
\paragraph{Datasets}
We consider the Kinetics-600 dataset~\cite{kay2017kinetics,carreira2018short} for class conditional video generation. 
Kinetics-600 is a large scale dataset of around 500k Youtube videos depicting 600 action classes. 
The videos are captured in the wild and exhibit lots of variability. 
The amount of videos available from Kinetics-600 is constantly changing as some videos become unavailable from streaming platforms.
We use a version of the dataset collected on June 2018.

Additionally, we use the BDD100K dataset~\cite{yu2018bdd100k} for unconditional video generation. 
BDD100K contains 100k videos recorded from inside cars representing more than 1000 hours of driving under different conditions. 
We use the training set split of 70K videos to train our models.

\paragraph{Evaluation metrics}
Defining proper evaluation metrics for video generation is an open research problem. 
We use metrics inspired by the image generation literature and adapted to video.
On Kinetics, we report three metrics: i) Inception Score (IS) given by an Inflated 3D Convnet~\cite{carreira2017quo} network trained on Kinetics-400, ii) Frechet Inception Distance on logits from the same I3D network, also known as Frechet Video Distance (FVD)~\cite{fvd}, and iii) Frechet Inception Distance on the last layer activations of an I3D network trained on Kinetics-600 (FID).
On BDD100K we report FVD and FID as described before, but we omit IS scores as they are not applicable since there are no labelled classes.

\begin{figure*}
    \centering
    \includegraphics[trim={0 80 0 0},clip,width=\textwidth]{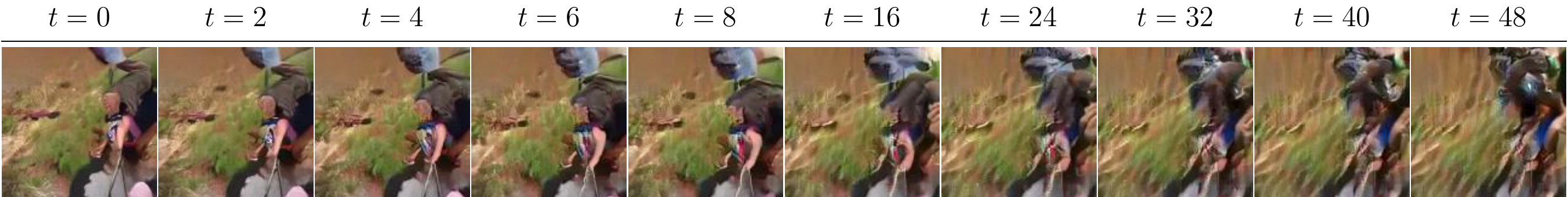}
    \caption{\small \textbf{DVD-GAN fails to generate samples beyond its training horizon}
    These samples were obtained by changing the spatial dimensions of the latent in a 6/128x128 \dvdgan model to produce 48/128x128 videos. The samples quickly degrade after the first few frames and become motionless.}
    \label{fig:dvdgan_unroll}
\end{figure*}

\begin{figure}
    \centering
    \includegraphics[trim={10 10 10 10},clip,width=0.85\linewidth]{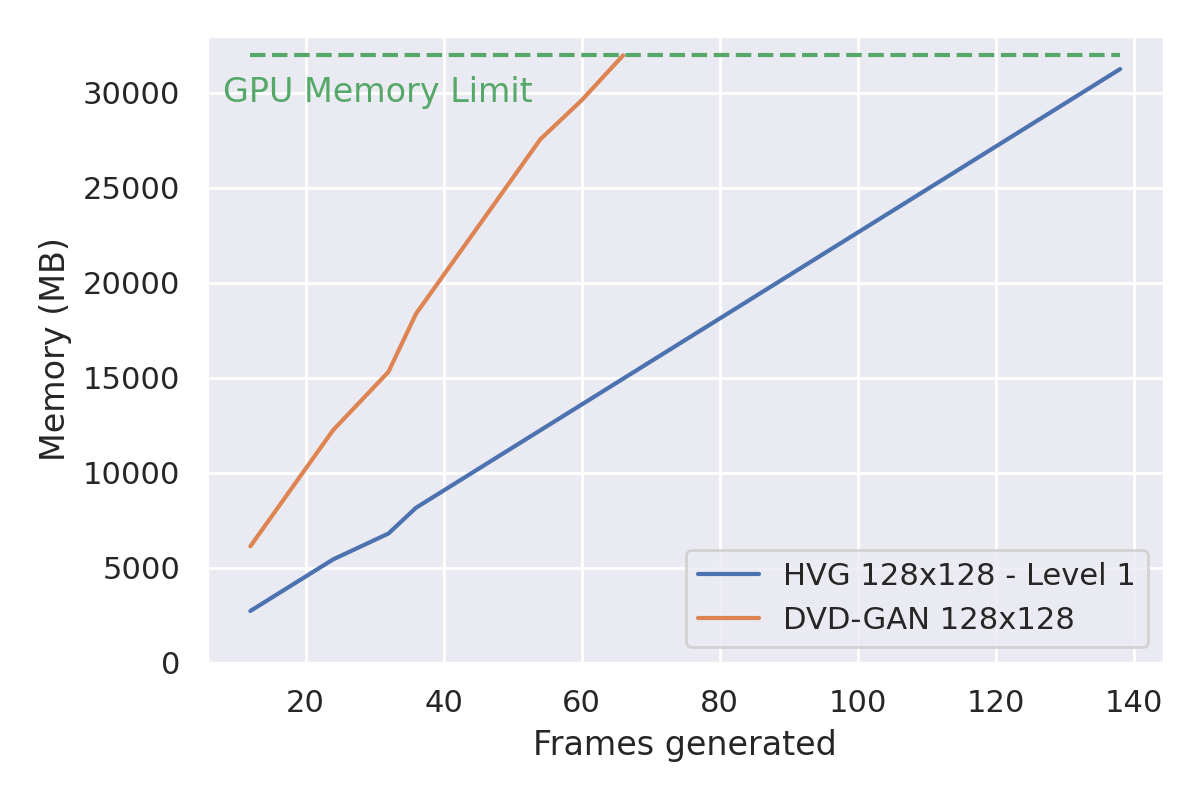}
    \caption{\small \textbf{Scaling the computational costs}
    This plot shows the required GPU memory for a two-level \ourmodel. We observe that the costs scales linearly with the output length for the first level, while the cost for the second level is fixed because it operates on a fixed length partial view of its input. Our model scales better than a comparable non-hierarchical model.}
    \label{fig:computational_costs}
\end{figure}

\paragraph{Implementation details}
All \ourmodel models are trained with a batch size of 512 examples and we fix a computational budget of up to 128 nVidia V100 GPUs, with most experiments requiring 64 GPUs.
Levels for Kinetics-600 are trained for 300k iterations, while levels for BDD100K are trained for 100k iterations.
We use early stopping if the metrics for a model stop improving.
We use the PyTorch framework and distribute training across multiple machines using data-parallelism. We use global-synchronous batch norm to synchronize the batch norm statistics across workers.

We employ the Adam~\cite{kingma2014adam} optimizer to train all levels with a fixed learning rate of \num{1e-4} for G and \num{5e-4} for D.
We use orthogonal matrices to initialize all the weights in our model as well as spectral norm in the generator and the discriminator.
Further implementation details can be found in the Appendix. 

\paragraph{Baselines}
As a baseline we consider \dvdgan~\cite{clark2019efficient}. 
DVD-GAN has shown state-of-the-art performance for video generation, being the first model to generate realistic samples on Kinetics-600 and surpassing MoCoGAN~\cite{tulyakov2018mocogan} and TGANv2~\cite{TGAN2017, saito2018tganv2} on simpler datasets.
We compare to DVD-GAN on Kinetics-600 generating 48/128x128 videos.

\subsection{Comparison to State-of-the-Art}
\label{sec:baseline_comparison}

We first investigate the performance obtained by our approach on Kinetics-600. 
We train a two-level \ourmodel on this dataset and compare it to \dvdgan which achieves, to our knowledge, the best generations on this dataset. 
We use the metrics reported in the original \dvdgan paper for 12/128x128 and 48/128x128 videos.\\
 
We train a two-level \ourmodel to generate 48/128x128 videos. 
The first level of the hierarchy generates 24/32x32 videos, with a temporal subsampling of 8 frames.
This level is therefore trained to generate videos of size 32 times lower than the final output size.
The second level upsamples the first level output using a factor of 2 for the temporal resolution and a factor 4 for the spatial resolution, producing 48/128x128 videos with a temporal subsampling of 4 frames.
During training the second stage operates on temporal crop of the first level generation. It takes as input windows of 6 frames and is trained to generate 12/128x128 video snippets, which are 4x lower dimensional than the final output.
As a result, this level has approximately the same training cost than a model generating videos of size 12/128x128.
At inference time we run the model convolutionally over all the 24 first level frames to generate 128x128 videos with 48 frames. 

To generate 12 frames with \ourmodel we use the same setup during training and inference.
Specifically, we run the first level to generate 24/32x32 frames, then we randomly select 6 of these frames, and finally we use these frames as the second level input to generate 12/128x128 videos.
To generate  48/128x128 videos, we apply the upsampling level convolutionally over the full first level output.

We report the scores obtain by our model in Table~\ref{tab:dvdgan_comparison}. For 12/128x128 videos, our model obtains higher IS and comparable FID to \dvdgan, validating that both models perform comparably when using a similar amount of computational resources.
Additionally, \ourmodel outperforms a 48/128x128 DVD-GAN model in FID score and reaches a similar IS score, despite only being trained on reduced views of the data.
Qualitatively, the generations of both models are similar - they do not degrade noticeable in appearance through time although both have some temporal inconsistencies. 
For \ourmodel temporal inconsistencies are usually the result of a bad generation from the first level.
For \dvdgan we hypothesize that most inconsistencies are due to the reduced temporal field of view of the discriminator, which is of less than 10 frames while the model generates 48 frames.

One of the main characteristics of \ourmodel is the ability to change the training and inference setup for upscaling levels.
Since \dvdgan is mostly a convolutional model, we investigate whether it can generate videos of longer duration than those it is trained on.
The number of frames in that model is controlled by the RNN that receives as input the latent variable sample and outputs as many tensors as frames to be generated.
We adjust the number of steps for this RNN and recompute batch normalization statistics to account for the extended amount of frames.
Some examples from an extended \dvdgan model can be found in the Appendix.
We observe that for this model videos are coherent up until the training length, but then they visibly degrade and become motionless, producing implausible generations.

\begin{figure*}
    \centering
    \includegraphics[trim={0 0 0 0}, clip, width=\textwidth]{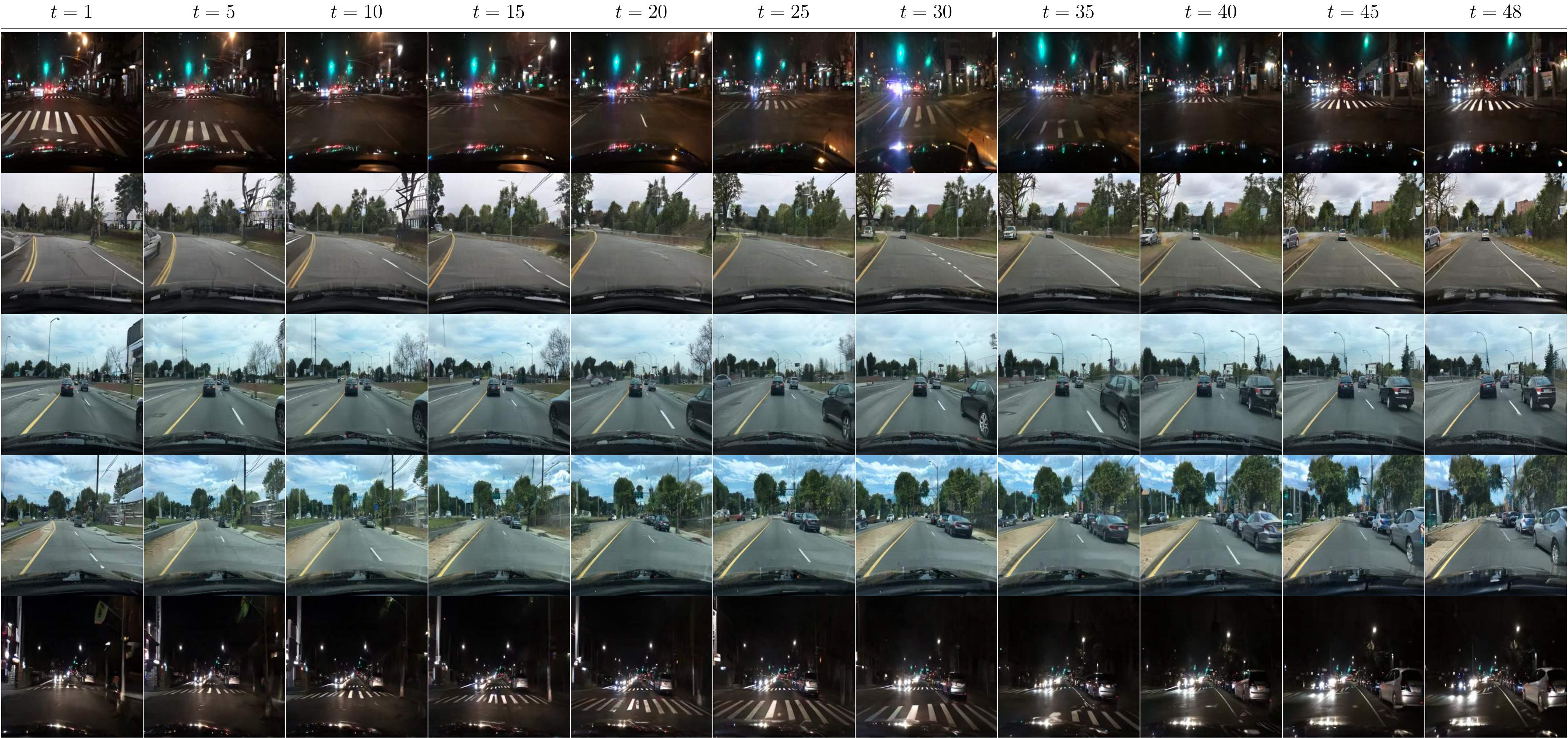}
    \caption{\small \textbf{Random 48/256x256 BDD100K samples:} We show samples from our three-stage BDD100K model. 
    Each row shows a different sample over time. 
    Despite the two stages of local upsampling, the frame quality does not degrade noticeably through time. 
    }
    \label{fig:samples_bdd_256}
\end{figure*}

\begin{figure*}
    \centering
    \includegraphics[trim={0 0 0 0},clip,width=\linewidth]{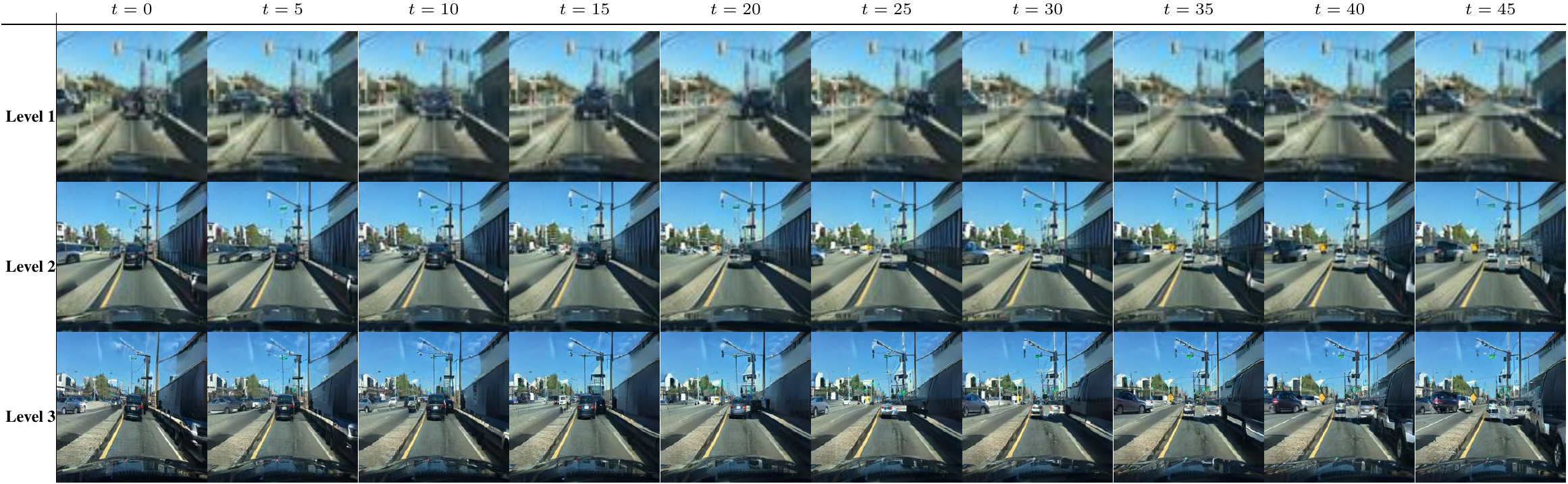}
    \caption{\small \textbf{\ourmodel failure mode}
    Here we show a prototypical low quality sample from our model. 
    We observe that upsampling levels add more details to the scene but do not produce a coherent generation, as the scene outline generated by the first level has some errors.
    We can trace back most failures to ambiguous generations from previous levels. 
    }
    \label{fig:bad_samples}
\end{figure*}

\subsection{Scaling-up \ourmodel}
\label{sec:scaling}
To show that our model scales well with the video dimensionality, we train a three-stage model on the BDD dataset to generate 48/256x256 videos.
A full training iteration for a batch with a single example of size 48/256x256 for a model similarly sized trained end-to-end requires more than 32GB of GPU memory - beyond the limits of commercial GPUs. 
Such a model is therefore not trainable on current GPUs, unless we use engineering techniques like gradient checkpointing or model parallelism which add a significant overhead to the training time.
Instead, we can train our model with a batch size of 512 using 128 GPUs.

We train the first level to output 12 frames at 64x64 resolution with a temporal subsampling of 8 frames.
The second level upsamples windows of 12 frames at 128x128 resolution with temporal subsampling of 4 frames (since we are doubling the framerate of the first level).
The third level is trained to upscale 12 frame windows at 256x256 resolution for a final temporal subsampling of 2 frames.
Figure~\ref{fig:samples_bdd_256} shows some samples from this model.
We observe a variety of scenes with different times (day/night) and different number of cars and objects.
The videos appear crisp and do not degrade through time.
However, we also observe some small temporal inconsistencies and wrong scenes for some generations.
An example of these kind of scenes is shown in Figure~\ref{fig:bad_samples}.
More samples from this model as well as its IS and FVD scores can be found in the Appendix.

To further illustrate the scaling capabilities of our model, we show the memory requirements for a two-level \ourmodel trained to generate 128x128 videos as a function of the number of output frames in Figure~\ref{fig:computational_costs}.
The first level generates half the total output frames at 64x64, while the second level is trained to upscale windows of 6 frames into 12 128x128 frames regardless of the first level output length.
Compared to a non-hierarchical model trained to generate 128x128 videos directly, our first level scales better thanks to the lower resolution and reduced number of frames.
Additionally, the second level has a fixed memory cost of 10290MB and is trained independently of the first stage. 
Given the same GPU memory budget, our model can generate sequences of up to 140 frames, more than twice the number of frames compared to a non-hierarchical model.
These computational costs are further increased when generating longer higher resolution videos with multiple upsampling stages.

\subsection{Model Ablations}
\label{sec:ablations}

\begin{figure*}[t]
    \centering
    \includegraphics[trim={0 0 0 0},clip,width=0.9\textwidth]{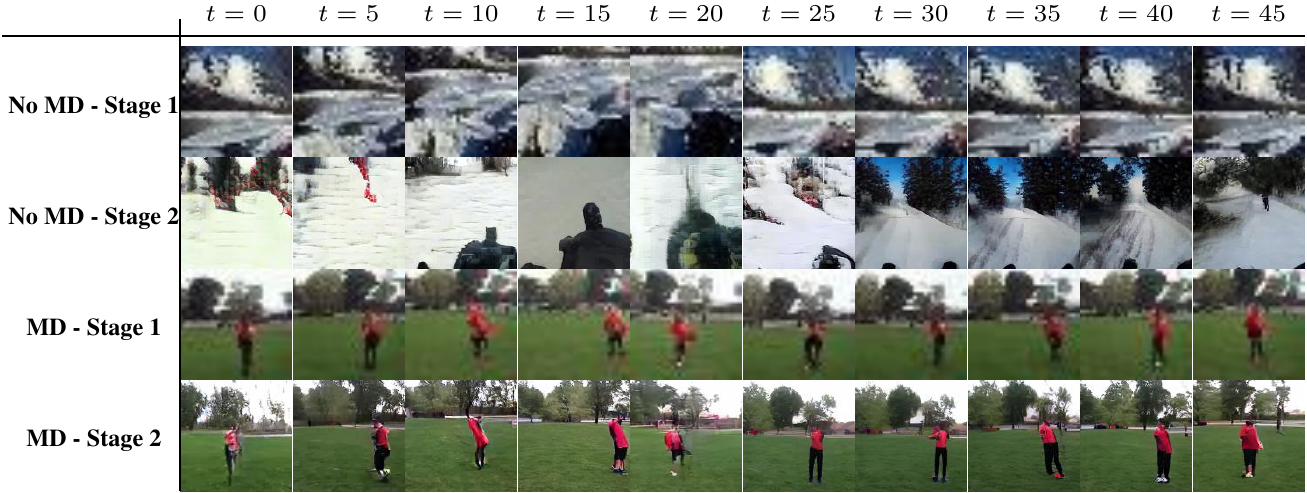}
    \caption{\small \textbf{Matching discriminator samples} We show a random sample from our two-level model on Kinetics-600 with the matching discriminator and without the matching discriminator (No MD). For each sample we show the output of the first level and the corresponding second level output. While the (no MD) model generates plausible local snippets at level 2, it does not remain temporally coherent. Our model with the matching discriminator is temporally consistent because it is grounded in the low resolution input.}
    \label{fig:md_samples}
\end{figure*}
\begingroup
\def\arraystretch{0.9}
\begin{table*}
\center
\begin{tabular}{cccccccc}
\toprule
& & \multicolumn{3}{c}{6 Frames} & \multicolumn{3}{c}{50 Frames} \\
\cmidrule(lr){3-5} \cmidrule(lr){6-8}
Dataset & Model & IS ($\uparrow$) & FID ($\downarrow$) & FVD ($\downarrow$) & IS ($\uparrow$) & FID ($\downarrow$) & FVD ($\downarrow$) \\
\midrule
  \multirow{2}{*}{Kinetics-600}  
& \ourmodel (No MD) & \textbf{50.31} & 1.62 & 594.99 & 37.81 & 42.29 & 1037.79 \\
& \ourmodel & 48.44 & \textbf{1.06} & \textbf{565.95} & \textbf{49.44} & \textbf{31.87} & \textbf{790.97} \\
\midrule
\multirow{2}{*}{BDD100K} & \ourmodel (No MD) & N/A & 1.36 & 211.69 & N/A & 26.52 & 575.51 \\
& \ourmodel & N/A & \textbf{1.07} & \textbf{144.96} & N/A & \textbf{18.73} & \textbf{326.78} \\
\bottomrule
\end{tabular}
\caption{\small \textbf{Matching discriminator comparison} 
We report metrics on Kinetics and BDD for our model with and without the matching discriminator. 
Both models perform similarly for 6 frames, corresponding to the training video length. 
However, the model without the matching discriminator does not perform well when applied over the full first level input because it is not grounded in its input.
}
\label{tab:results_kinetics}
\end{table*}
\endgroup

\paragraph{Matching Discriminator Ablation}
To assess the importance of the matching discriminator, we compare two-level \ourmodel models with and without the matching discriminator (we refer to the latter as No MD). 
We use the same setup as in Section~\ref{sec:baseline_comparison}, where we generate 48 frames with a two-level model.
We train the second level on 3-frame windows of the first level to generate 6 frames.

We expect the No MD model to fail at generating coherent full-length videos when applied over the full first level output since its outputs are not necessarily grounded on its inputs.
On 6 frame generations (i.e.\ the. training setup), \ourmodel and \ourmodel No MD obtain similar scores as reported in Table~\ref{tab:results_kinetics}. 
While the No MD model ignores its previous level inputs, it still learns to generate plausible 6 frame videos at 128x128.
However, when we use the models to generate full-length 48 frame videos, we observe that \ourmodel No MD generates valid local snippets but is inconsistent through time.

Fig.~\ref{fig:md_samples} shows an example of a full length No MD generation in which this effect is clearly observable.
In contrast, our model with a matching discriminator stays grounded and consistent through time.
This is reflected in the reported metrics in Table~\ref{tab:results_kinetics}, where the No MD model obtains significantly worse scores. 
This ablation highlights the importance of using a matching discriminator, as it ensures that upsampling level outputs are grounded to its inputs.

\begingroup
\begin{table}
\center
\begin{tabular}{cccc}
\toprule
& \multicolumn{3}{c}{48 Frames} \\
\cmidrule(lr){2-4}
Model & IS ($\uparrow$) & FID ($\downarrow$) & FVD ($\downarrow$) \\
\midrule
3-frame  & 58.21 & 31.59 & 714.74 \\
6-frame  & \textbf{77.36} & \textbf{14.00} & \textbf{517.21} \\
\bottomrule
\end{tabular}
\caption{\small \textbf{Temporal Window Ablation} 
We analyze the impact of the temporal window for upsampling levels by comparing models trained with different window sizes.
The models are trained with 3-frame or 6-frame windows for the upsampling level.
While the 6-frame model has higher computational requirements, it outperforms the 3-frame model, confirming that there is a trade-off between computational savings and final performance when selecting the size of the temporal window for upsampling stages. 
}
\label{tab:window_ablation}
\end{table}
\endgroup

\paragraph{Temporal Window Ablation}
One of the modelling choices in \ourmodel is the temporal window length used in the upsampling levels.
Shorter inputs provide less context to upsample frames, while longer inputs require more compute.
To assess the impact of the window length, we compare two-level models trained on Kinetics-600 128x128: one trained on first level windows of 6 frames (same setup as in Section~\ref{sec:baseline_comparison}) and one trained on windows of only 3 frames.
The 6-frame level requires approximately 2x GPU memory than the 3-frame level during training, but we expect it to perform better due to the larger context available for upscaling.
We compare their performance to generate 48 frames in Table~\ref{tab:window_ablation}.

We conclude that the window size defines a trade-off between computational resources - shorter temporal windows require less computation - and sample quality - longer windows have more context to fill in details when upsampling.

%% file: conclusions.tex
\section{Conclusions}

We propose \ourmodel, a hierarchical video generator that divides the generative process into multiple simpler steps.
Our model is competitive with state-of-the-art approaches in terms of sample quality, while
enabling higher resolutions generations for longer temporal horizons than possible before.
Higher capacity models that produce larger outputs are key aspects for improving video generation, and \ourmodel is a step in that direction that has better scaling properties than previous approaches.

%% file: appendix.tex
\clearpage
\appendix

\section{Additional Implementation Details}
We use the Adam optimizer~\cite{kingma2014adam} with learning rate $\lambda_G = 1e10^{-4}$ and $\lambda_D = 5e10^{-4}$ for the generator and the discriminator, respectively. The discriminator is updated twice for each generator update.

We use orthogonal initialization for all the weights in our model and use spectral normalization both in the generator and the discriminator.
We only use the first singular value to normalize the weights.
We do not use weight moving averages nor orthogonal penalties. %

Conditional batch normalization layers use the input noise as the condition, concatenated with the  class label when applicable.
Features are normalized with a per-frame mean and standard deviation.

To unroll a generator beyond its training temporal horizon, we apply it convolutionally over longer input sequences.
We perform 200 ``dummy'' forward passes to recompute the per-timestep batch normalization statistics at test time.

All convolutions in our models use 3x3 or 3x3x3 filters with padding=1 and stride=1, for 2D or 3D convolutions respectively.
All models were implemented in PyTorch.

\section{Model Architecture Details}
For the rest of the section, we use B to denote the batch size, T for the number of frames or timesteps, C for the number of channels, H for the height of the frame and W is the width of the frame.

\subsection{First level architecture}

\paragraph{Generator}
The generator is composed by a stack of units where each unit is comprised of a ConvGRU layer and two 2D-ResNet upsampling blocks. 
We follow the nomenclature of~\cite{brock2018large, clark2019efficient} and describe our network using a base number of channels $ch$ and the channel multipliers associated with each unit. 
Our first level generator is formed by 4 units with channel multipliers $[8, 8, 4, 2]$. 
The base number of channel is $128$.

The first input of this network is of size BxTx(8x$ch$)x4x4.
This input is obtained by first embedding the class label onto a 128 dimensional space, then concatenating the embedding to a 128 dimensional noise vector. 
This concatenation is mapped to a Bx(8x$ch$)x4x4 tensor with a linear layer and a reshape, and then the final tensor is obtained by replicating the output of the linear layer T times.

The ConvGRU layer~\cite{ballas2015delving} follows the ConvGRU implementation of ~\cite{clark2019efficient} and uses a ReLU non-linearity to compute the ConvGRU update.

The 2D ResNet blocks are of the norm-act-conv-norm-act-conv style.
We use conditional batch normalization layers, ReLU activations and standard 2D convolutions.
Before the first convolution operation and after the first normalization and activation, there is an optional upsampling operation when increasing the resolution of the tensor.
We use standard nearest neighbor upsampling. 
Except for the last unit, all units perform this upsampling operation.
The conditional batch normalization layers receive the embedded class label (if applicable) and the input noise as a condition and map it to the corresponding gain and bias term of the normalization layer using a learned linear transformation.
The 2D ResNet blocks process all frames independently by reshaping their input to be (B*T)xCxHxW.

The output of the last stack goes through a final norm-relu-conv-tanh block that maps the output tensor to RGB space with values in the [-1, 1] range.

\paragraph{Discriminator}
There are two discriminators, a 2D spatial discriminator and a 3D temporal discriminator.
The 2D discriminator is composed of 2D ResNet blocks.
Each ResNet block is formed by a sequence of relu-conv-relu-conv layers.
There are no normalization layers in the discriminator.
After the last conv in each block there is an optional downsampling operation, which is implemented with average pooling layers.
The 2D discriminator receives as input 8 randomly sampled frames from real or generated samples.

The 3D discriminator is equal to the 2D discriminator except that its first two layers are 3D ResNet blocks, implemented by replacing 2D convolutions with regular 3D convolutions.
The 3D discriminator receives as input a spatially downsampled (by a factor of two) real or generated sample.
The 2D blocks process different timesteps independently.

We concatenate the output of both discriminators and use a geometric hinge loss.
The loss is averaged over samples and outputs.

We use $128$ as base number of channel for both discriminators, with the following channel multipliers for each ResNet block: $[16, 16, 8, 4, 2]$

\subsection{Upsampling levels architecture}
The upsampling levels follow the same architecture as the first level but with the following modifications.

\paragraph{Generator}
The generator units replace the ConvGRU layers with a Separable 3D convolution. 
We first convolve over the temporal dimension with a 1D temporal kernel of size 3 and then convolve over the spatial dimension with 2D 3x3 kernel. 
We empirically compare generators with ConvGRU and separable convolutions in section~\ref{app:comparison}, showing that the 3D convolution performs as well as the ConvGRU but it can be run in parallel.

We add residual connections at the end of each 3D and 2D ResNet block to an appropriately resized version of $\mathbf{x}^{l-1}$.
We use nearest neighbor spatial downsampling for this operation, and we use nearest neighbor temporal interpolation to increase the number of frames of $\mathbf{x}^{l-1}$. We then map the residual to the appropriate number of channels using a linear $1$x$1$ convolution.
We do not add $\mathbf{x}^{l-1}$ residual connections to feature maps with spatial resolutions (HxW) greater than the resolution of $\mathbf{x}^{l-1}$.

\paragraph{Discriminator}
We reuse the same 2D and 3D discriminators as for the first stage.
Additionally, we add a matching discriminator that discriminates $(\mathbf{x}^{l}, \mathbf{x}^{l-1})$ pairs.
The matching discriminator utilizes the same architecture as the 3D discriminator.
It receives as input a concatenation of $\mathbf{x}^{l-1}$ and a downsampled version of $\mathbf{x}^{l}$ to match the resolution of $\mathbf{x}^{l-1}$.
We concatenate the outputs of all three networks and use a geometric hinge loss, as done for the first level discriminator.
The overall loss is averaged over samples and output locations.

For 128x128 generations on Kinetics, the generator uses $128$ as base number of filters with the following channel multipliers $[8, 8, 4, 2, 1]$. All discriminators have $96$ base channels and the following channel multipliers $[1, 2, 4, 8, 16, 16]$.
All our Kinetics models at 128x128 are two-level models.
We train models to upsample inputs of sizes 3/32x32 or 6/32x32 to 6/128x128 or 12/128x128, respectively.
Since we train our first level for 24/32x32 outputs, our two-level models can generate 48/128x128 outputs when unrolled.

For 128x128 generations on BDD100K, the generator uses $96$ as the base number of channels with channel multipliers $[8, 8, 4, 2, 1]$. 
All discriminators have $96$ base channels and channel multipliers $[1, 2, 4, 8, 16, 16]$.
Our BDD100K 128x128 models upsample 6/64x64 inputs to 12/128x128, and can generate outputs of up to 24/128x128. 

For 256x256 generations on BDD100K, the generator uses $96$ as the base number of channels with channel multipliers $[8, 4, 4, 4, 2, 1]$. 
All discriminators have $96$ base channels and channel multipliers $[1, 2, 4, 8, 8, 16, 16]$.
Our 256x256 model upsamples 6/128x128 inputs to 12/256x256, and can generate outputs of up to 48/256x256.

\section{Comparison of Recurrent Layers}
\label{app:comparison}

\begin{figure}
\begin{subfigure}{0.45\textwidth}
    \centering
    \includegraphics[width=\textwidth]{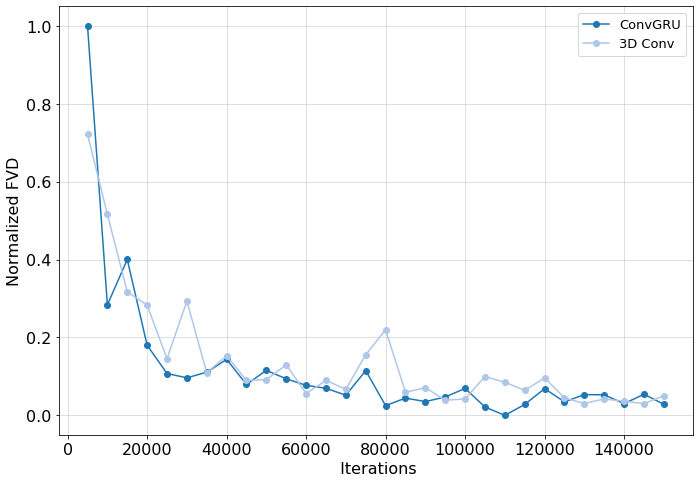}
\end{subfigure}
\hfill
\begin{subfigure}{0.45\textwidth}
    \centering
    \includegraphics[width=\textwidth]{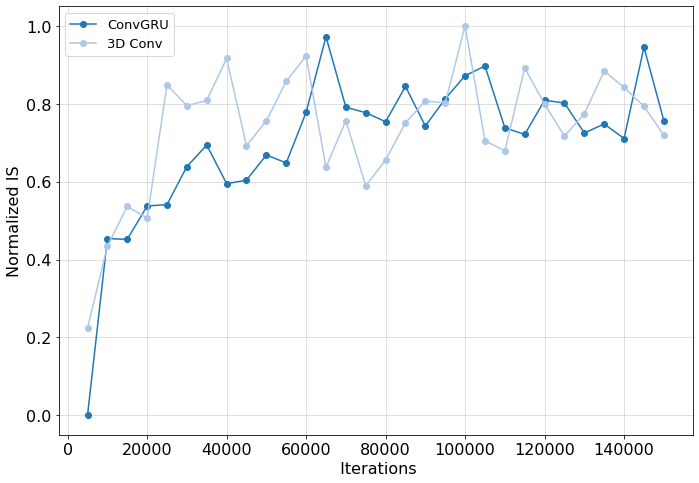}
\end{subfigure}
\caption{\textbf{Comparison of recurrent layers} We compare two variants of the same generator, one with a single ConvGRU layer per generator block and one with a separable 3D convolution per generator block. 
On the left we show the evolution of the FVD score during training, and on the right we show the Inception Score.
Both scores are normalized to the [0, 1] range where 1 is the highest score obtained by these models and 0 the lowest.
Both models have similar behaviour and computational costs, but the 3D convolution processes inputs in parallel.}
\label{fig:appendix_conv3d}
\end{figure}

In this section we justify the change of the ConvGRU for Separable 3D convolutions in upsampling levels.
In Figure~\ref{fig:appendix_conv3d} we compare the evolution of two metrics (IS and FVD) during training for two variants of the same two-stage model, one using ConvGRUs and one using separable 3D convolutions.
Both models show similar behavior during training and achieve similar final metrics.
However, ConvGRUs perform sequential operations over time whereas 3D convolutions can be parallelized.

\clearpage
\section{Power Spectrum Density}

\begin{figure}[!b]
    \centering
    \includegraphics[trim={10 10 10 10}, clip, width=0.95\linewidth]{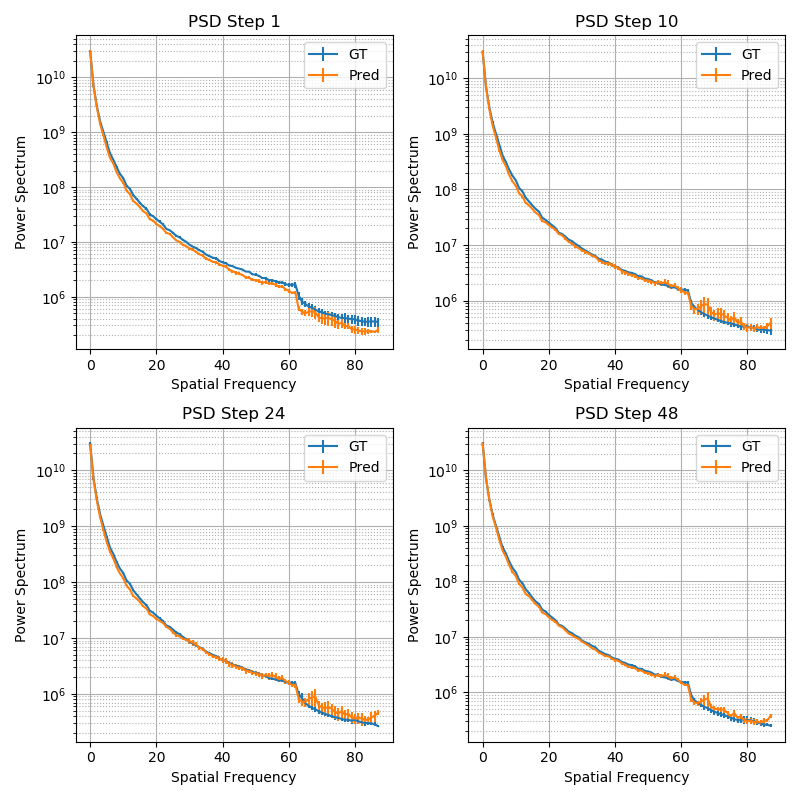}
    \caption{\textbf{Power Spectrum Density (PSD) plots for different time steps} We show a comparison of the PSD between the original data and our generations at different steps in the predictions. We observe that our generations have a similar PSD to that of the original data, even at the end of the generation, indicating that the generations do not blur over time significantly.}
    \label{fig:psd}
\end{figure}

Some video generation models produce blurry results over time. 
As an additional evaluation, we generate Power Spectrum Density (PSD) plots to assess whether our generations become blurrier over time, following~\cite{ayzel2020rainnet}.

We conduct this experiment on Kinetics for videos of 48 frames at 128x128 resolution.
We use our model with the first level trained on 24/32x32 sequences and the second level trained to generate 12/128x128 video snippets from 6/32x32 windows, and unrolled after training to produce 48/128x128 videos.
We took 1800 random videos from the ground truth data (GT) and 1800 generations from our model.
We compute the PSD at frames 1, 10, 24, and 48 of each video.
For each set of 600 videos, we compute the average PSD across videos, on a per frame basis.
Finally, we use the three sets of 600 videos to compute the standard deviation and mean for the average PSD of the original data and our generations.
Figure~\ref{fig:psd} shows the plots for different frame indices. 
Our generations have a very similar PSD to that of GT in all video frames.
This indicates that our generations, while they might not be accurate, have very similar frequency statistics as the ground-truth data.
We do not observe any significant blurring over time, which is confirmed by the plots - they show that even for frame 48 high frequencies are very similar between the original data and our generations.

\section{Influence of Motion on the Results}

\begin{table}[ht]
    \centering
    \begin{tabular}{clccc}
    \toprule
    & & \multicolumn{2}{c}{48 frames} \\
    \cmidrule(lr){3-4} 
    & Class & IS ($\uparrow$) & FID  ($\downarrow$) & \# Videos \\
    \midrule
          \parbox[t]{2mm}{\multirow{5}{*}{\rotatebox[origin=c]{90}{\textbf{High Motion}}}} & Bungee Jumping & 11.66 & 82.21  & 799 \\
          & Capoeira       & 9.48  & 84.57  & 816 \\
          & Cheerleading   & 12.10 & 116.84 & 982 \\
          & Kitesurfing    & 11.36 & 108.81 & 648 \\
          & Skydiving      & 5.99  & 90.25  & 983 \\
    \midrule
          \parbox[t]{2mm}{\multirow{5}{*}{\rotatebox[origin=c]{90}{\textbf{Low Motion}}}} & Doing Nails  & 13.32 & 91.67  & 537 \\
          & Cooking Egg  & 7.60  & 111.63 & 441 \\
          & Crying       & 8.92  & 70.74  & 627 \\
          & Reading Book & 11.13 & 64.97  & 793 \\
          & Yawning      & 9.71. & 79.08  & 530 \\
    \bottomrule
    \end{tabular}
    \caption{\small \textbf{Per category scores for classes with different amounts of motion (Kinetics-600)} We report per-class IS and FID scores for 5 randomly selected categories with high motion and 5 categories with low motion. We observe that there is a high variability in FID scores, with some classes with low motion having high scores as well as some high motion classes. In IS scores there are few differences between the two groups, with the high motion group having a slightly higher mean score.}
    \label{tab:motion_metrics}
\end{table}

In this section we analyze whether our model has different performance for categories with different motion characteristics as an additional analysis. 

We conduct this experiment for \ourmodel trained on Kinetics-600 to generate 24/32x32 videos in the first level and then to upscale 6/32x326 videos to 12/128x128 videos for the second level. 
The second level is unrolled over the full first level generation to obtain 48/128x128 videos.

We randomly select five categories of videos with high motion content (bungee jumping, capoeira, cheerleading, kitesurfing and skydiving) and five categories with less dynamic videos (doing nails, cooking egg, crying, reading book and yawning).
We generate 1000 samples from \ourmodel for each category, and use all available samples in the dataset to compute IS and FID scores per category.

\begin{figure*}
    \centering
    \includegraphics[width=1.0\textwidth]{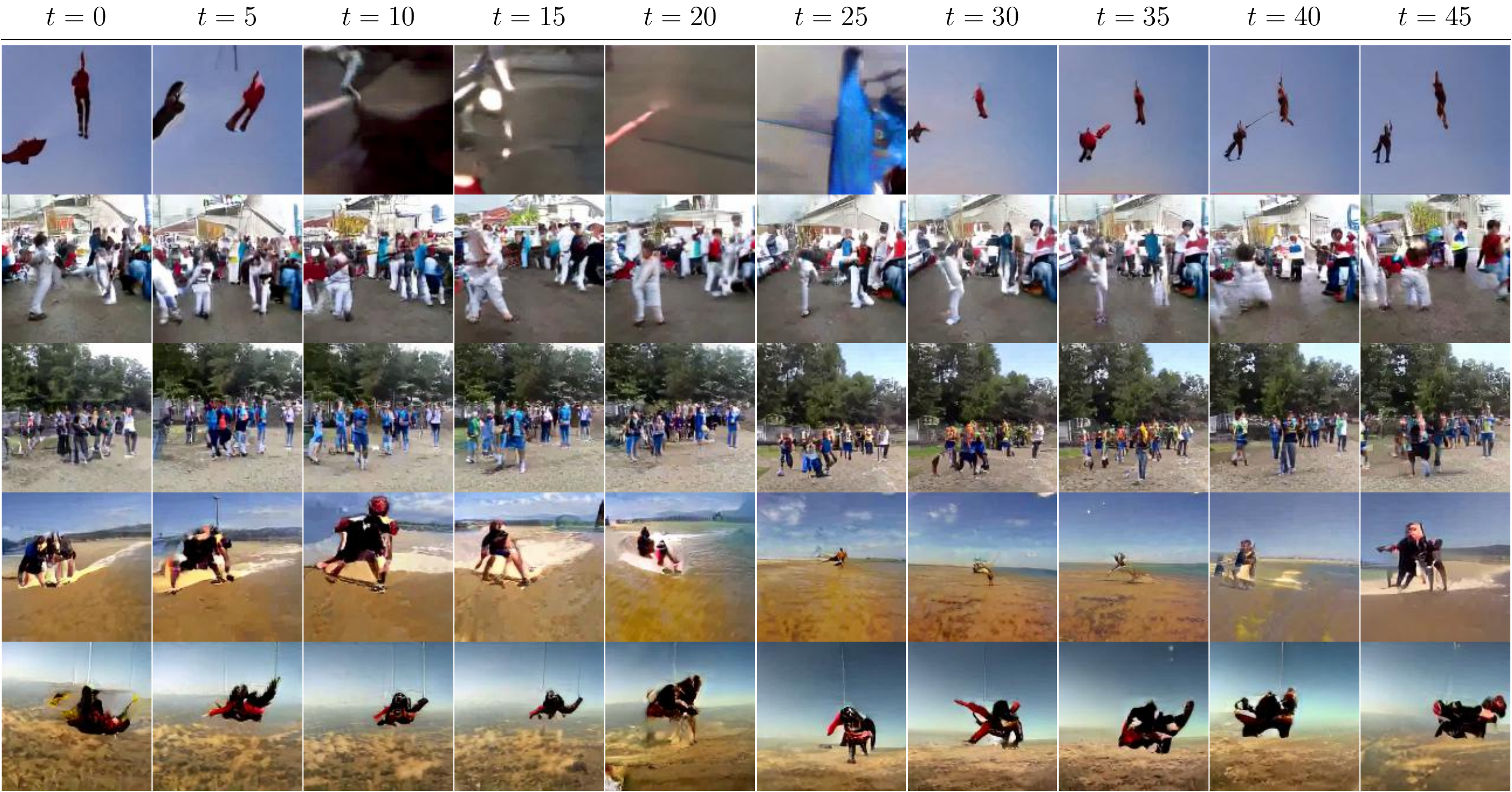}
    \caption{\textbf{Samples from Kinetics-600 classes with high motion content}}
    \label{fig:samples_high_motion}
\end{figure*}

\begin{figure*}
    \centering
    \includegraphics[width=1.0\textwidth]{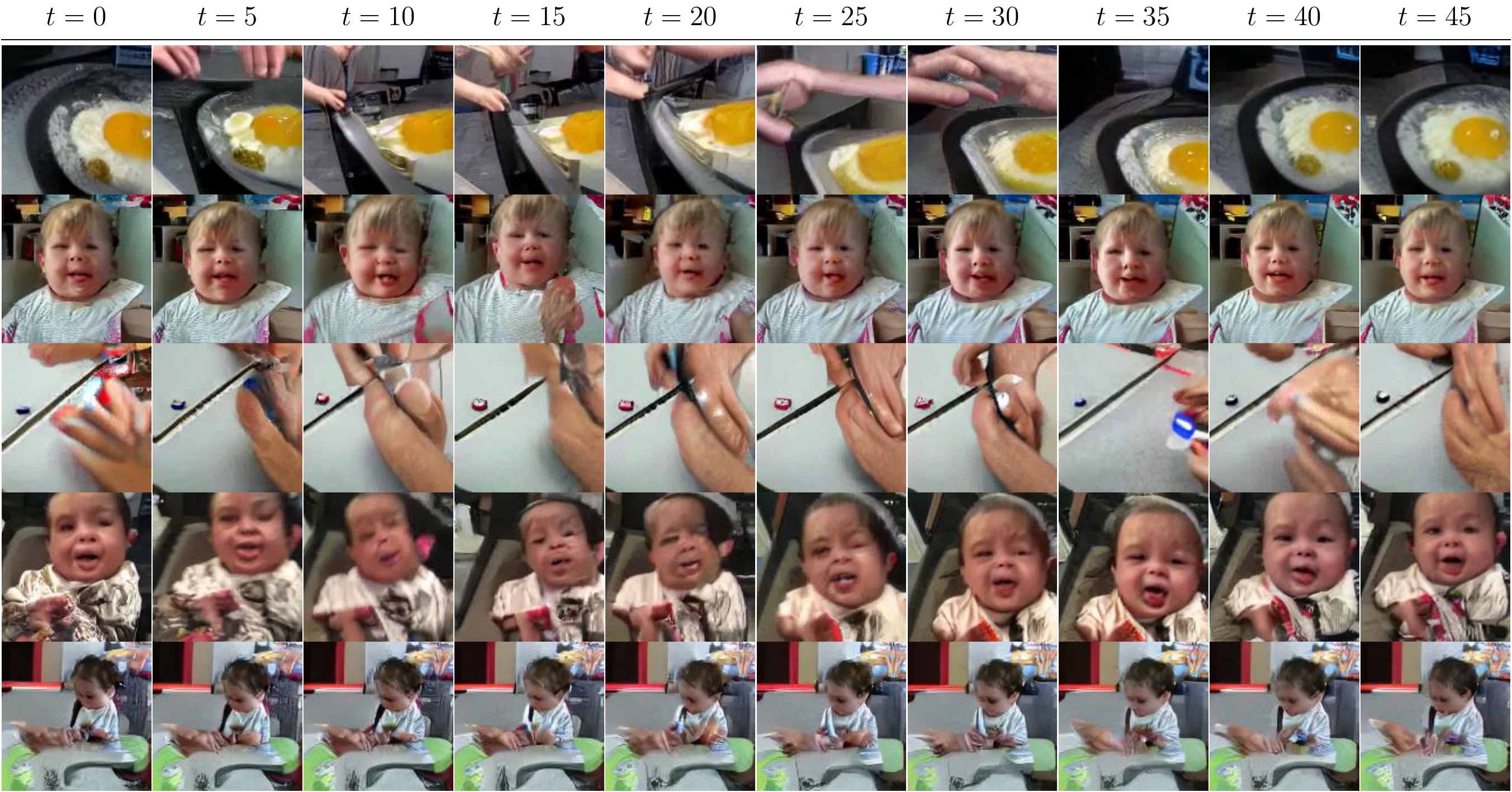}
    \caption{\textbf{Samples from Kinetics-600 classes with low motion content}}
    \label{fig:samples_low_motion}
\end{figure*}

Table~\ref{tab:motion_metrics} shows the IS and FID scores of each class, while Figure~\ref{fig:samples_high_motion} and Figure~\ref{fig:samples_low_motion} show samples from high motion and low motion categories, respectively.
We do not observe a trend that indicates that \ourmodel produces worse generations for high motion classes. 
Some low motion categories have high FID scores similar to the highest scores for the high motion categories, while on average the IS scores for the high motion categories are slightly better.
We do not notice a qualitative difference.
Instead, we believe there might be other factors - amount of structure present in a scene for example -  that have greater impact on the output quality.

\begin{table*}[!h]
    \centering
    \begin{tabular}{cccccc}
    \toprule
    & & \multicolumn{2}{c}{Evaluated on 12 frames} & \multicolumn{2}{c}{Evaluated on 48 frames} \\
    \cmidrule(lr){3-4} \cmidrule(lr){5-6} 
    Model & Trained on & FID ($\downarrow$) & FVD ($\downarrow$)  & FID ($\downarrow$) & FVD ($\downarrow$) \\
    \midrule
    3-Level \ourmodel & 12/256x256 & 3.66 & 541.37 & 21.38 & 391.69 \\ 
    \bottomrule
    \end{tabular}
    \caption{\small \textbf{BDD100K 256x256 Metrics} We report the FID and FVD scores for our three-level \ourmodel trained on BDD100K. The model is trained to generate 12/256x256 videos and at inference it produces 48/256x256 videos.}
    \label{tab:bdd_256_metrics}
\end{table*}

\section{Hierarchical Training Objective}
In this section we describe our training objective more formally.
For a \ourmodel model with $L$ levels, our goal is to model the joint probability distribution $p_{d}(\mathbf{x}^1, ..., \mathbf{x}^L) = p_{g}(\mathbf{x}^1, ..., \mathbf{x}^L) = p_{g_L}(\mathbf{x}^L | \mathbf{x}^{L-1}) .. p_{g_1}(x^1)$, where each $p_{g_l}$ is defined by a level $l$ in our model.

\paragraph{Training Level 1} We consider the distribution $p_{g_1}$ and solve a min-max game with the following value function:
\begin{eqnarray*}
& V_1(G_1, D_1) = \\
& \mathbb{E}_{\mathbf{x}^1\sim p_{d}} [\log (D_1(\mathbf{x}^1))] + \mathbb{E}_{\mathbf{z}_1 \sim p_{z_1}} [\log (1-D_1(G_1(\mathbf{z}_1)))],
\end{eqnarray*} where $G_1$ and $D_1$ are the generator/discriminator associated with the first stage and $p_{z_1}$ is a noise distribution.

This is the standard GAN objective. As shown in~\cite{goodfellow2014generative}, the min-max game $\min_{G_1} \max_{D_1} V_1(G_1, D_1)$ has a global minimum when $p_{g_1}(\mathbf{x}^1) = p_{d}(\mathbf{x}^1)$.

\paragraph{Training upsampling levels} 
For each upscaling level $l > 1$ we formulate a min-max game with the following value function:
{\small
\begin{eqnarray*}
& V_l(G_l, D_l) = \\
&  \mathbb{E}_{\mathbf{x}^{l-1}, ..., \mathbf{x}^{1}\sim p_{d}} \mathbb{E}_{\mathbf{x}^l\sim p_{d}( .| \mathbf{x}^{l-1}, ..., \mathbf{x}^{1})} [\log(D_l(\mathbf{x}^{l}, \mathbf{x}^{l-1}))] + \nonumber \\ & \mathbb{E}_{\mathbf{\hat{x}}^{l-1}\sim p_{g_{l-1}}} \mathbb{E}_{\mathbf{z}_l\sim p_{z_l}} [\log(1-D_l(G_l(\mathbf{z}_l, \mathbf{\hat{x}}^{l-1}),  \mathbf{\hat{x}}^{l-1}))],
\end{eqnarray*}} where $G_l$, $D_l$ are the generator and discriminator of the current level and $p_{g_{l-1}}$ is the generative distribution of the level $l-1$.

The min-max game $\min_{G_l} \max_{D_l} V_l(G_l, D_l)$ has a global minimum when the two joint distributions are equal, $p_{d}(\mathbf{x}, ..., \mathbf{x}^l) = p_{g_l}(\mathbf{x}^l | \mathbf{x}^{l-1}) .. p_{g_1}(\mathbf{x}^1)$~\cite{dumoulin2016adversarially, donahue2016adversarial}. 
It follows that $p_{d}(\mathbf{x}^l | \mathbf{x}^{l-1}) = p_{g_l}(\mathbf{x}^l | \mathbf{x}^{l-1})$ when $p_{g_{l-1}}(\mathbf{x}^{l-1} | \mathbf{x}^{l-2}) .. p_{g_1}(\mathbf{x}^1) = p_{d}(\mathbf{x}^{l-1}, ..., \mathbf{x}^1)$.
Level $l$ only learns the parameters associated with the distribution $p_{g_l}$, as all $p_{g_i}, 1 \leq i \leq l-1 $ levels are trained previously and their training objectives admit a global minimum when they match the data distribution. 
However, even if the distribution $p_{g_{l-1}}$ does not match exactly the marginal data distribution, our model still aims at learning a distribution $p_{g_l}$ such that $p_{g_l}(\mathbf{x}^l | \mathbf{x}^{l-1}).. p_{g_1}(\mathbf{x}^1)$ approximates the joint data distribution.

\section{BDD100K metrics}
In this section we report the metrics for our three-level BDD100K model for help future comparison to \ourmodel.
Metrics are shown in Table~\ref{tab:bdd_256_metrics}.

\section{Additional Samples}
Additional samples can be found in .mp4/.gif format along with this appendix in the supplementary materials file.
These videos show multiple samples from our two-level model for Kinetics-600 and our three-level model for BDD100K. 

We have included 3 videos. 
One video shows samples from our BDD100K \ourmodel model. 
For each sample it shows the output of each of the three levels, with the output of the last level being 48/256x256. 
Another video shows samples from our 12/128x128 model unrolled to generate 48/128x128 samples on Kinetics-600.
Finally, we include a gif file showing random samples from a \dvdgan model that produces outputs beyond its training horizon.
This video shows that all samples from this model diverge.

To complement the videos, we also include some additional samples for our Kinetics-600 48/128x128 model and BDD100K 48/256x256 model below.

For all evaluations, we sample from an isotropic Gaussian with unit variance for ease of comparison and reproducibility.
Samples for figures and the provided video files were produced by sampling with standard deviation $\sigma = 0.5$.
We observed that noise samples with reduced variance produce higher quality samples but are slightly less diverse.

\paragraph{Qualitative comparison with \dvdgan}
As a point of comparison we provide samples from the official \dvdgan 48/128x128 trained on Kinetics-600 and released by the authors.
Samples can be download in this \href{https://drive.google.com/file/d/1P8SsWEGP6tEGPPNPH-iVycOlN6vpIgE8/view?usp=sharing}{link} . 
We observe that the samples from \dvdgan and our \ourmodel model are of similar quality.

\begin{figure*}
    \centering
    \includegraphics[trim={0 80 0 0}, clip,width=0.95\linewidth]{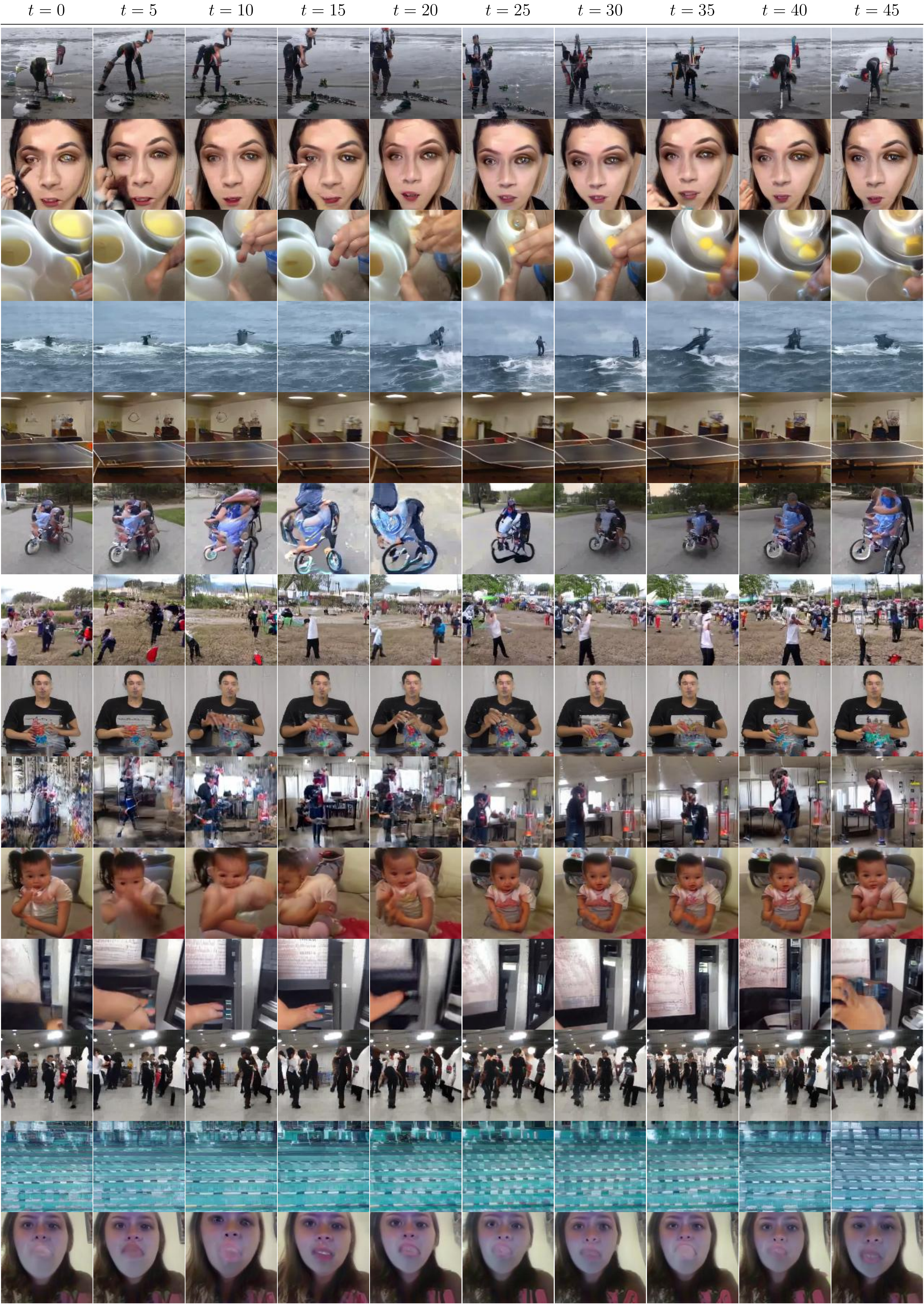}
    \caption{\textbf{Additional samples for Kinetics 12/128x128} We show additional samples from our two-level Kinetics 12/128x128 model unrolled to generate 48/128x128 videos. More samples can be found in the supplementary videos.}
    \label{fig:samples_kinetics_appendix}
\end{figure*}

\begin{figure*}
    \centering
    \includegraphics[width=\linewidth]{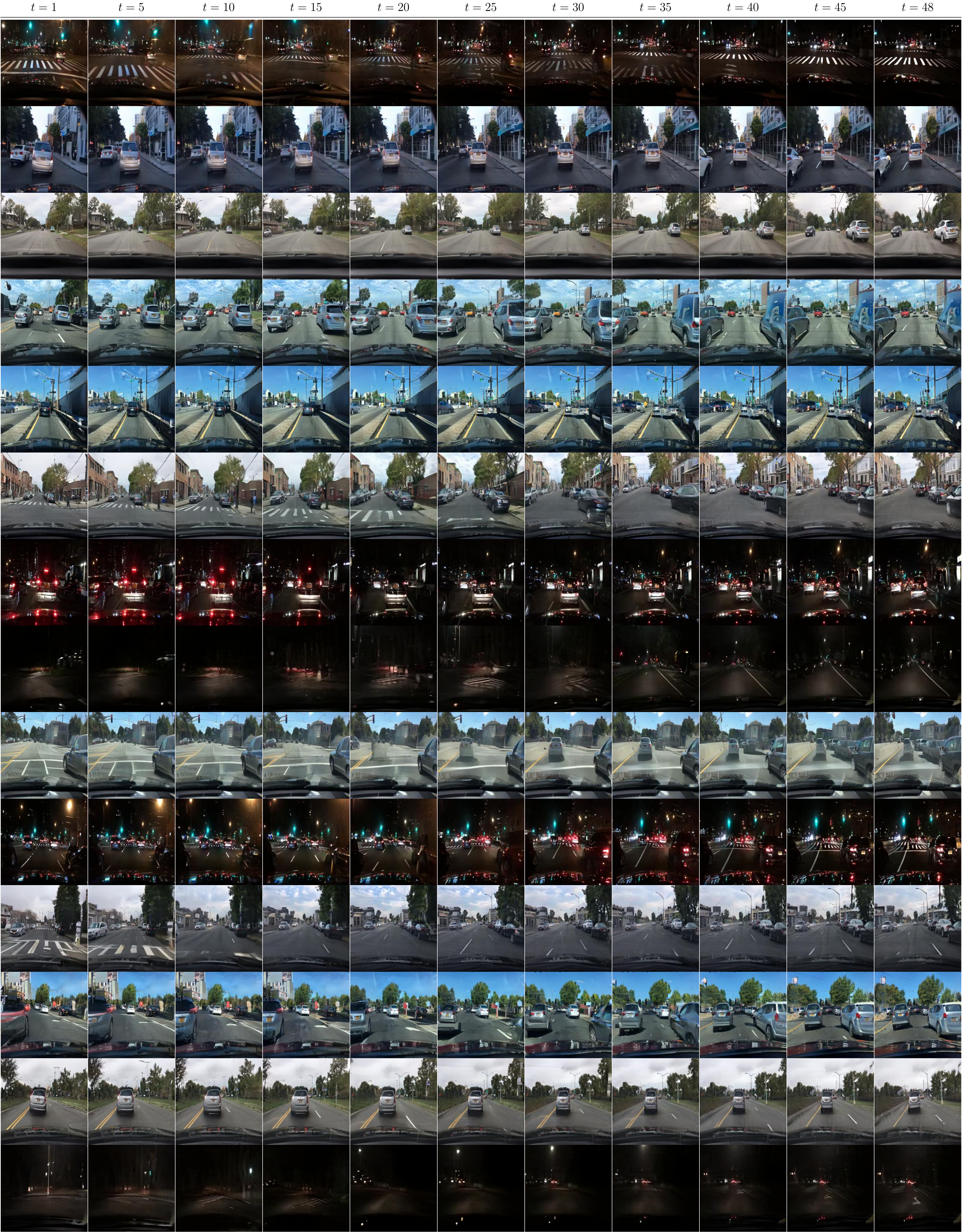}
    \caption{\textbf{Additional samples for BDD 48/256x256} We show additional samples from our three-level BDD 48/256x256 model.}
    \label{fig:samples_bdd_appendix}
\end{figure*}

\section{Upsampling Visualizations}
\begin{figure*}[ht]
    \centering
    \includegraphics[trim={0 65 0 0}, clip, width=0.85\textwidth]{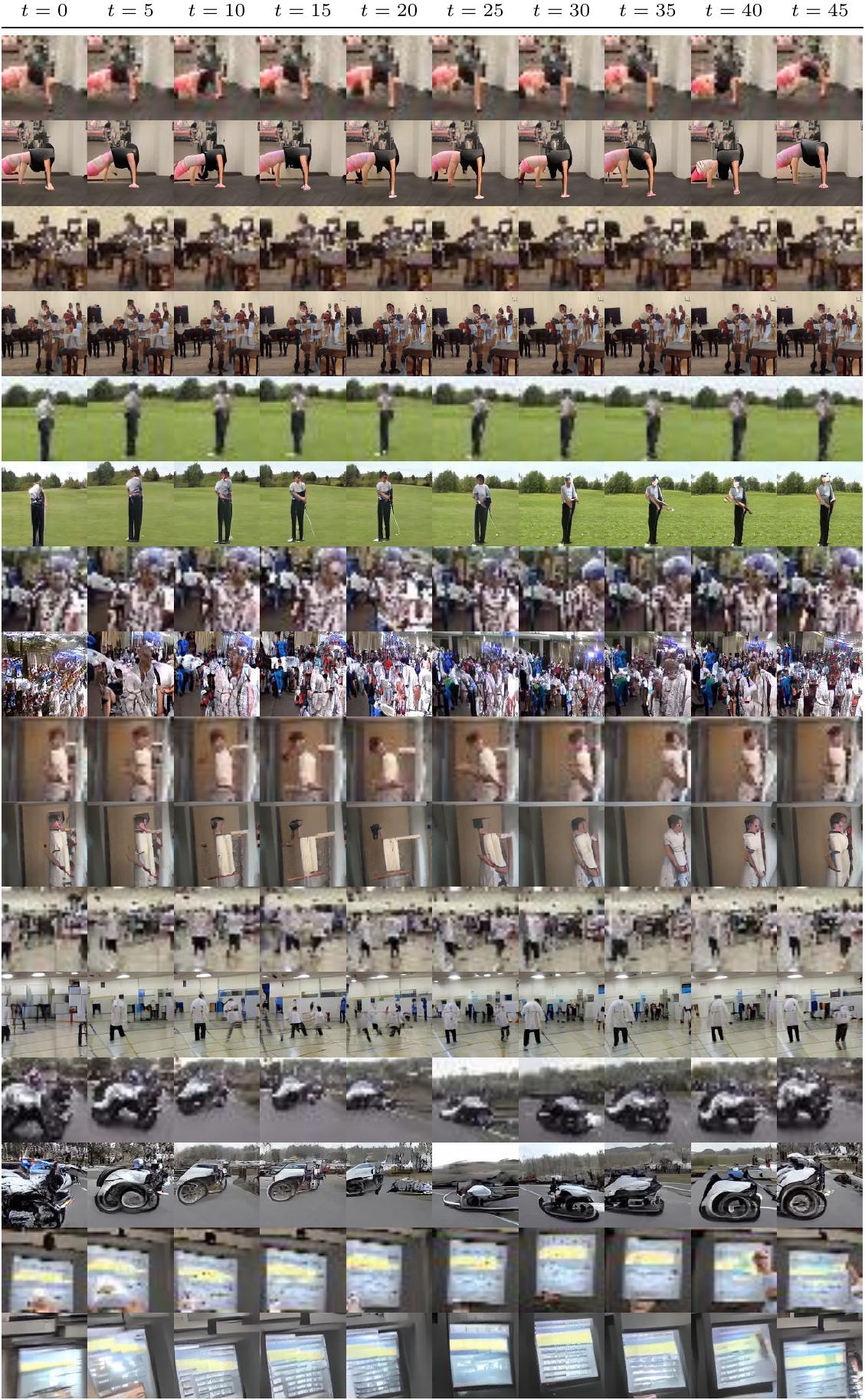}
    \caption{\textbf{Pairs of samples from stage 1 and their corresponding stage 2 output} We show a few examples from our 12/128x128 two-level model trained on Kinetics-600 and unrolled to generate 48/128x128 videos. For each example, we show the first level low resolution generation and the corresponding level 2 upsampling. Level 2 outputs refine the details of the first level generations but retain the overall scene structure.}
    \label{fig:appendix_upsampling}
\end{figure*}

In this section we show some examples of level 1 generations on Kinetics-600 for a 24/32x32 model, as well as the corresponding 48/128x128 generations from level 2 trained to upscale 6/32x32 windows to 12/128x128 and unrolled over the whole first level generation.
Examples are shown in Figure~\ref{fig:appendix_upsampling}, in which, for each example, we show the level 1 generation on the top row and the corresponding level 2 generation in the lower row.
We observe that the second level adds details and refines the low resolution generation beyond simple upsampling, but at the same time keeps the overall structure of the low resolution generation and is properly grounded.